\documentclass[conference]{IEEEtran}
\IEEEoverridecommandlockouts

\usepackage{multirow}
\usepackage{multicol}
\usepackage{cite}
\usepackage{amsmath,amssymb,amsfonts}
\usepackage{algorithm}
\usepackage{algpseudocode}

\usepackage{graphicx}
\usepackage{textcomp}
\usepackage{xcolor}

\makeatletter
\newcommand{\linebreakand}{%
\end{@IEEEauthorhalign}
\hfill\mbox{}\par
\mbox{}\hfill\begin{@IEEEauthorhalign}
}
\makeatother

\DeclareMathOperator*{\argmax}{arg\,max}
\DeclareMathOperator*{\argmin}{arg\,min}

\def\BibTeX{{\rm B\kern-.05em{\sc i\kern-.025em b}\kern-.08em
    T\kern-.1667em\lower.7ex\hbox{E}\kern-.125emX}}
\begin{document}

\title{Mitigating Long-Tailed Anomaly Score Distributions with Importance-Weighted Loss\\
{}
}

\author{


\IEEEauthorblockN{Jungi Lee}
\IEEEauthorblockA{\textit{R\&D Division} \\
\textit{ELROILAB Inc.}\\
Seoul, Republic of Korea \\
ganbbang12@elroilab.com}
\and
\IEEEauthorblockN{Jungkwon Kim}
\IEEEauthorblockA{\textit{R\&D Division} \\
\textit{ELROILAB Inc.}\\
Seoul, Republic of Korea \\
jkkim@elroilab.com}
\and

\IEEEauthorblockN{Chi Zhang}
\IEEEauthorblockA{\textit{R\&D Division} \\
\textit{ELROILAB Inc.}\\
Seoul, Republic of Korea \\
czhang@elroilab.com}

\linebreakand

\IEEEauthorblockN{Sangmin Kim}
\IEEEauthorblockA{\textit{R\&D Division} \\
\textit{ELROILAB Inc.}\\
Seoul, Republic of Korea \\
smkim@elroilab.com}

\and

\IEEEauthorblockN{Kwangsun Yoo}
\IEEEauthorblockA{\textit{R\&D Division} \\
\textit{ELROILAB Inc.}\\
Seoul, Republic of Korea \\
yks@elroilab.com}

\and

\IEEEauthorblockN{Seok-Joo Byun\textsuperscript{*}}
\IEEEauthorblockA{\textit{R\&D Division} \\
\textit{ELROILAB Inc.}\\
Seoul, Republic of Korea \\
sjbyun@elroilab.com}

}

\maketitle
\begingroup\renewcommand\thefootnote{*}
\footnotetext{Corresponding Author}

\begin{abstract}
Anomaly detection is crucial in industrial applications for identifying rare and unseen patterns to ensure system reliability. Traditional models, trained on a single class of normal data, struggle with real-world distributions where normal data exhibit diverse patterns, leading to class imbalance and long-tailed anomaly score distributions (LTD). This imbalance skews model training and degrades detection performance, especially for minority instances. To address this issue, we propose a novel importance-weighted loss designed specifically for anomaly detection. Compared to the previous method for LTD in classification, our method does not require prior knowledge of normal data classes. Instead, we introduce a weighted loss function that incorporates importance sampling to align the distribution of anomaly scores with a target Gaussian, ensuring a balanced representation of normal data. Extensive experiments on three benchmark image datasets and three real-world hyperspectral imaging datasets demonstrate the robustness of our approach in mitigating LTD-induced bias. Our method improves anomaly detection performance by 0.043, highlighting its effectiveness in real-world applications.
\end{abstract}


\section{Introduction}

Anomaly detection is a field dedicated to identifying unseen data as anomalies. Typically, anomaly detection models are trained only on a single class of normal data, as anomalous data are often scarce or unavailable. However, in real-world industrial settings, normal data frequently comprise a diverse set of objects or patterns rather than a single homogeneous category. This inherent diversity can introduce class imbalance within the normal data, resulting in a right-skewed anomaly score distribution, commonly known as a Long-Tailed Distribution (LTD). For instance, in hyperspectral anomaly detection, datasets~\cite{963e-1d34-24} often consist of background data and a diverse set of normal objects. Fig.~\ref{fig:1} visualizes this phenomenon using t-distributed Stochastic Neighbor Embedding (t-SNE)\cite{hinton2002stochastic}, illustrating both the data and anomaly score distributions. The visualization reveals that the data form multiple clusters rather than a single compact cluster, leading to a right-skewed anomaly score distribution. A key issue arises from the overlapping variance between the tail class (minority) and head class (majority), which biases the model toward the majority class during training. As a result, the model’s ability to detect anomalies within the tail class degrades significantly. This challenge underscores the need for advanced methodologies that effectively address class imbalance and enhance the robustness of anomaly detection models in complex industrial environments.

In classification tasks, numerous approaches have been developed to tackle the challenges posed by LTD. These methods generally fall into three broad categories~\cite{zhang2023deep}: class re-balancing, information augmentation, and model enhancement. Class re-balancing aims to mitigate class imbalance by adjusting the distribution of training data. Common techniques include under-sampling~\cite{liu2008exploratory} or over-sampling based on class frequency~\cite{chawla2002smote}, as well as employing weighted loss functions that inversely scale with the frequency of each class~\cite{zhang2021learning}. Data augmentation techniques, such as mix-up~\cite{zhang2018mixup}, have been proposed to alleviate the effects of LTD~\cite{zhong2021improving}. These methods generate synthetic samples to improve the representation of minority classes. Model enhancement approaches involve improving the model architecture or training strategy. For instance, contrastive learning strengthens feature representations by leveraging relations between data points~\cite{yang2020rethinking}. However, these methods are primarily designed for multi-class classification, where prior knowledge of multiple class labels is available. In anomaly detection, this assumption is often impractical, as the task typically involves only a single class of normal data and lacks explicit multi-class information. As a result, directly applying these techniques to anomaly detection poses challenges and requires careful adaptation to accommodate the unique constraints and characteristics of the domain.

\begin{figure}[t]
  \centering
  \includegraphics[width=0.45\linewidth]{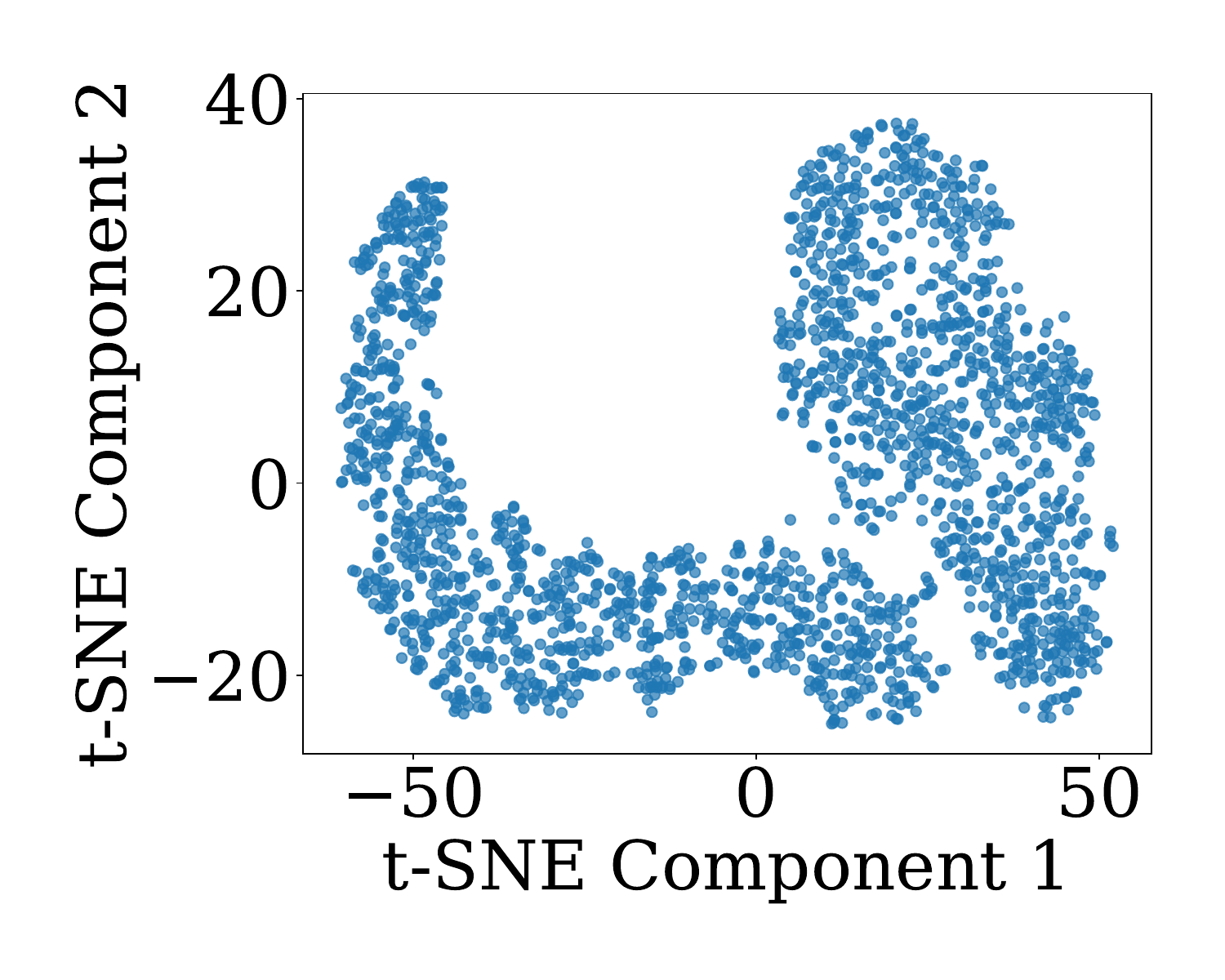}
  \includegraphics[width=0.45\linewidth]{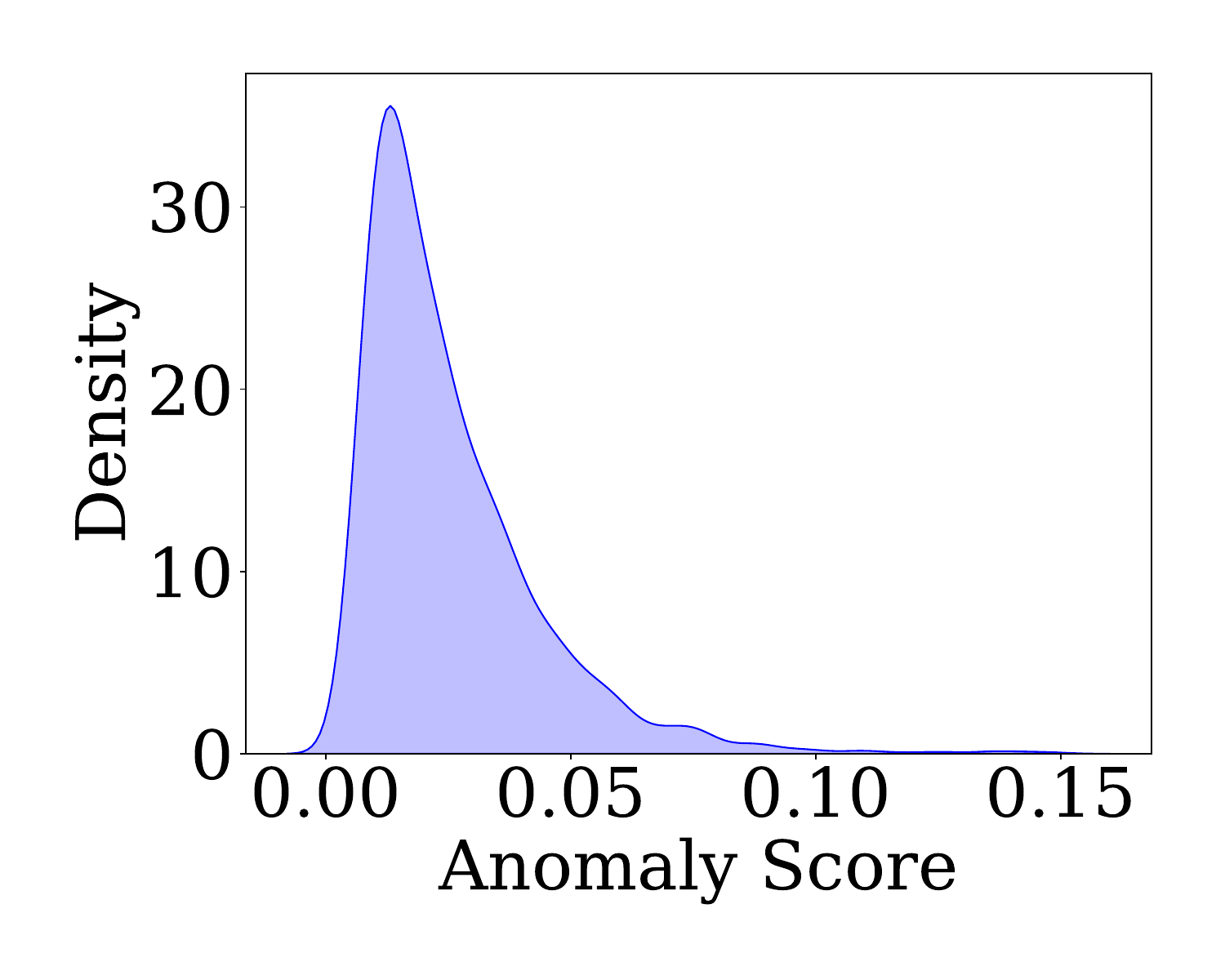}
  \caption{Analysis of the hyperspectral anomaly detection dataset~\cite{963e-1d34-24}. We analyze the hyperspectral anomaly detection dataset and visualize the data using t-SNE~\cite{hinton2002stochastic} (left). The distribution of anomaly scores (right) exhibits a long-tailed pattern, highlighting the inherent class imbalance and its impact on anomaly detection.}
  \label{fig:1}
\end{figure}

In anomaly detection, the presence of diverse classes within normal data often results in a LTD. Unlike classification problems, where LTD solutions are well-established, addressing this issue in anomaly detection is significantly more challenging due to the absence of explicit multi-class information. To mitigate this, some approaches employ generative models to augment tail-class instances, aiming to balance the anomaly score distribution~\cite{kim2020gan}. Additionally, \cite{kim2020gan} proposes a method that utilizes k-means clustering to sample from imbalanced datasets. However, these methods face several limitations, such as the absence of a clear definition for identifying tail instances, the constraints of k-means clustering, and difficulties in scaling to large datasets. These challenges highlight the need for more adaptive and scalable solutions to effectively address LTD in anomaly detection.

To overcome these challenges, we propose a novel Importance Weighted Loss (IWL) that operates without requiring prior knowledge of the number or nature of normal data classes. Our approach is founded on the assumption that the anomaly score distribution should follow a Gaussian distribution. To enforce this, we utilize importance sampling weights, which are derived from the discrepancy between the current anomaly score distribution and the target Gaussian distribution. This adaptive weighting ensures that the model effectively balances tail-class and head-class instances, mitigating the bias introduced by long-tailed distributions. By aligning the anomaly score distribution with a target Gaussian, our method improves the representation of normal data and enhances anomaly detection performance, even in highly imbalanced datasets.

We provide the following contributions:
\begin{itemize}
    \item We design a loss function for anomaly detection specifically tailored for long-tailed distributions, eliminating the need for prior knowledge of the number or characteristics of normal data classes.
    \item We propose an importance-weighted loss, which leverages importance sampling based on the anomaly score distribution, ensuring that the distribution aligns with a Gaussian form for improved anomaly detection performance.
    \item We validate the necessity and effectiveness of our proposed loss function through extensive experiments on three image datasets and real-world hyperspectral image datasets, demonstrating its superiority in handling long-tailed distributions in anomaly detection.
\end{itemize}

\section{Related work}
To address the long-tailed distribution problem, representative approaches include under-sampling and over-sampling. One method is to randomly select head-class instances to match the number of tail-class instances, or to replicate tail-class instances to match the head-class count~\cite{liu2008exploratory, estabrooks2004multiple}. However, these methods can lead to underfitting of the head class or overfitting due to excessive replication of tail-class instances. Cost-sensitive learning re-balances by assigning different loss values to each class—penalizing the head class or assigning more weight to the tail class to improve generalization performance~\cite{elkan2001foundations,zhou2005training,sun2007cost}. In transfer learning, head-to-tail knowledge transfer is used to transfer information from the head class to the tail class~\cite{wang2017learning, wang2021rsg}. Recently, many studies have utilized contrastive learning~\cite{yang2020rethinking} and two-stage training~\cite{Kang2020Decoupling} in which the feature extractor and the downstream task model are retrained. However, these methods either require prior knowledge of class information or are applicable only to tabular data, and thus cannot be used in the anomaly detection domain, which typically consists of a single class.

One approach to addressing class imbalance in anomaly detection is to assign weights to the tailed data during training. However, this method leads to an overfitting of the minority class that forms the long-tailed distribution, thereby degrading performance~\cite{ding2019modeling}. Sampling with $k$-means cluster~\cite{kim2020gan} proposed a sampling strategy for the imbalance problem, whereas it relies on the number of $k$ and degrades the performance with a few sub-classes in a normal class. Data augmentation with generative models can address the scare of long-tailed data. In the long-tailed distribution, the generative models rather degrade the performance of the minority class. Reference~\cite{Ho_2024_CVPR} addresses the issues by employing a text-derived feature vector that requires additional models. Previous approaches require extensive research or the use of new models, making it difficult to integrate them with conventional anomaly detection models.

\section{Preliminary}
Consider a dataset $X = \{ x_i \}_{i=1}^n \subset \mathbb{R}^{n \times m}$, where $n$ is the number of data and $m$ is the dimension of the data. When training an autoencoder (AE) $f(\cdot)$, we minimize an error $\| e_i \|_2^2$, where $e_i = x_i-f(x_i) \in \mathbb{R}^{m}$.

\textbf{Distribution of outputs}
Minimizing the mean-squared error loss (MSE) $\| e_i \|_2^2$ is equivalent to the maximum likelihood optimization problem, which can be stated as follows:
\begin{equation}\label{e1}
        \argmin_\theta \sum^m_{j=1}||x_{ij} - f(x_{ij})||^2_2 = \argmax_{\theta}\prod^m_{j=1}P(f(x_{ij})|x_{ij},\theta).
\end{equation}
When the model is trained sufficiently well, the maximum likelihood in the above approximates the output distribution such that 
\begin{equation}
    \max_{\theta}\prod^m_{j=1}P(f(x_{ij})|x_{ij},\theta) = P(Y|X) \approx P(Y).
\end{equation}
In \cite{ding2019modeling}, the authors observed that the optimization problem in \eqref{e1} can be interpreted through the lens of Bregman's theory~\cite{banerjee2005clustering}, leading to an optimized empirical distribution that is effectively expressed via a Gaussian kernel.
In particular, by applying kernel density estimation, the optimized distribution $\hat{P}(Y)$ of $P(Y)$ is given by
\begin{equation}
    \hat{P}(Y) = \frac1n \sum_{j=1}^m \mathcal{N}(f(x_{ij}), \sigma^2),
\end{equation}
which provides a smooth estimation of the underlying distribution.

\textbf{Distribution of anomaly scores}
Leveraging the aforementioned relation, we can derive the distribution of anomaly scores.
Indeed, the residual error between $x$ and its reconstruction $f(x)$ follows a Gaussian distribution, i.e., 
\begin{equation}
    e_i=x_i-f(x_i) =\bigl(e_{i1}, e_{i2}, \cdots, e_{im}\bigr) \sim \mathcal{N}(0,\sigma^2).
\end{equation}
The anomaly score $s_i$ for data $x_i$ is defined by the squared Euclidean norm of $e_i$ as
\begin{equation}
    s_i = ||x_i - f(x_i)||_2^2 = \sum_{j=1}^{m} e_{ij}^2.    
\end{equation}
Hence, we focus on the distribution of $e_{ij}$ to figure out that of $s_i$.
Since each $e_{ij} \sim \mathcal{N}(0, \sigma^2)$ is independent for $1 \leq j \leq m$, the squared term $e_{ij}^2$ follows a scaled chi-square distribution $\sigma^2\chi^2_1$.    
Combining this and the assumption that each $e_{ij}$ is independent of the others, the anomaly score satisfies
\begin{equation}
    \frac{s_i}{\sigma^2} = \frac{1}{\sigma^2} \sum_{j=1}^{m} e_{ij}^2 \sim \chi^2_m, 
\end{equation}
where $\chi_m^2$ is a chi-squared distribution with $m$ degrees of freedom. Thus, the distribution of the anomaly score $s_i$ can be characterized as $\sigma^2 \cdot \chi^2_m$.    
When a random variable $Z \sim \chi_m^2$, it is well known that $\mathbb{E}(Z) = m$ and $\operatorname{Var}(Z) = 2m$. 
Hence, we see 
\begin{equation}
    \mathbb{E}[s_i] = \sigma^2 m \quad \text{and} \quad \operatorname{Var}(s_i) = 2 \sigma^4 m.
\end{equation}
By the Central Limit Theorem (CLT), the anomaly score, expressed as a form of  summation
\begin{equation}
    s_i = \sum_{j=1}^{m} e_{ij}^2
\end{equation}
converges to a normal distribution for sufficiently large number of features $m$.
As a result, the distribution of anomaly scores with MSE loss function follows a Gaussian distribution. 

\section{Long-tailed Distribution Loss}
We demonstrated that the anomaly score distribution is expected to follow a Gaussian distribution. However, as long-tailed distributions arise in classification problems due to class imbalances, anomaly detection also exhibits a long-tailed distribution as a result of data imbalance within the normal class. Fig.~\ref{fig:2} illustrates the anomaly score distribution obtained from training on the MNIST dataset, where classes 6 and 7 are designated as normal data. The ``balanced distribution'' represents an equal proportion of data from both classes, while the ``imbalanced distribution'' results from defining class 6 as the majority class and class 7 as the minority class. Due to this imbalance, the skewness of the log anomaly score distribution increases from 0.504 to 0.524, demonstrating that the presence of a minority class induces a long-tailed distribution. Using conventional MSE loss, it is challenging to effectively learn from long-tailed data. Moreover, assigning high weights to tail-class instances during training can lead to overfitting~\cite{ding2019modeling}. Overfitting occurs because the model focuses excessively on the minority class, neglecting the majority class. This suggests that properly adjusting weights can balance overfitting and underfitting, mitigating the challenges posed by long-tailed distributions. Since the overfitting problem arises from a distorted distribution dominated by the minority class, we propose a weighted loss function based on importance sampling with a normal distribution. This method adjusts weights dynamically, ensuring a more balanced learning process while effectively addressing the limitations of long-tailed distributions in anomaly detection.

\begin{figure}[t]
  \centering
  \includegraphics[width=0.7\linewidth]{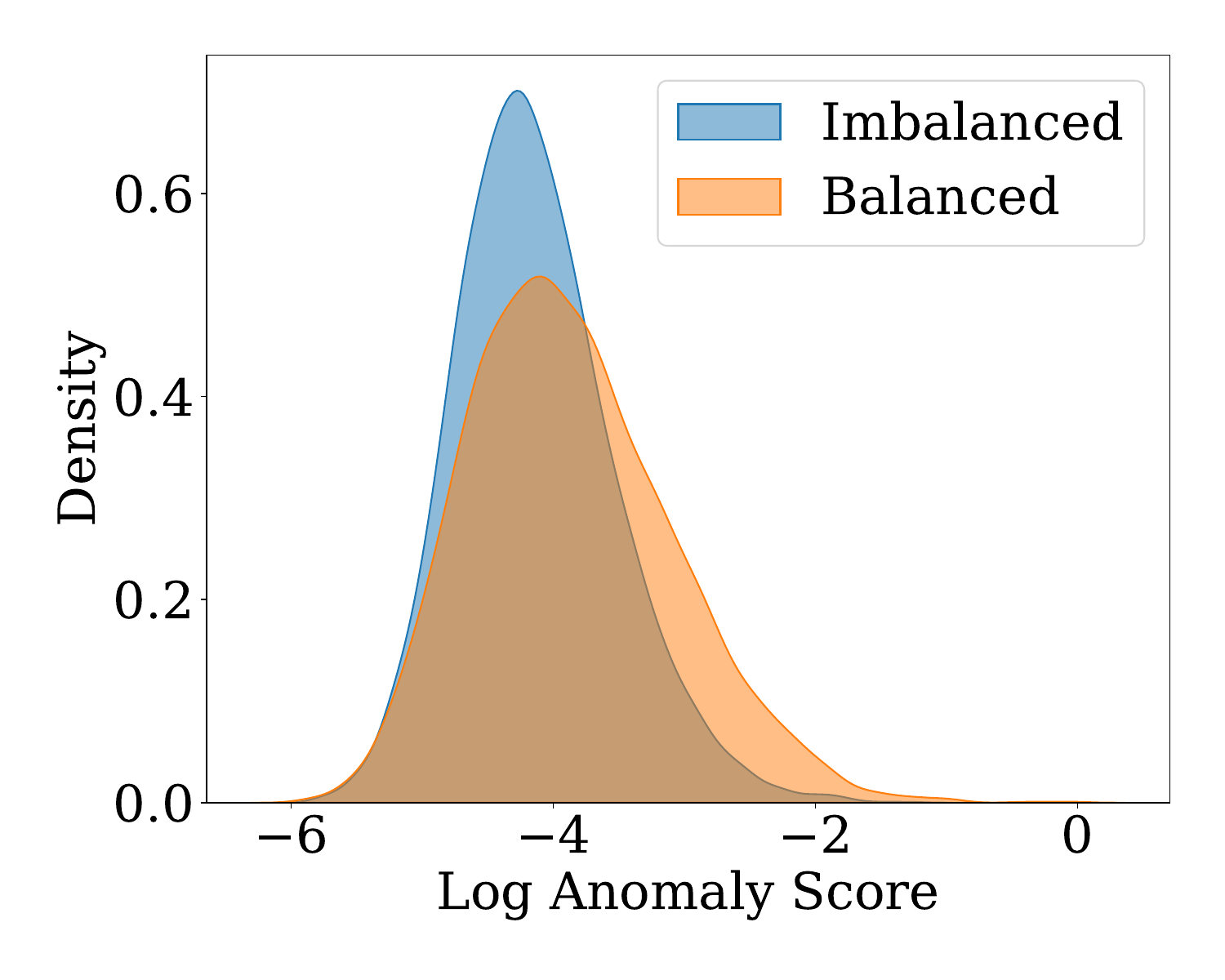}
  \caption{Distribution of log anomaly scores in the MNIST dataset. We designate classes ``6'' and ``7'' as the majority and minority classes, respectively. To construct a balanced dataset, both classes are assigned an equal number of samples. Conversely, in the imbalanced dataset, the minority class comprises only one-tenth of the samples in the majority class. The introduction of class imbalance leads to an increase in skewness from 0.504 to 0.524, indicating that class imbalance inherently induces a long-tailed distribution in anomaly score distributions.}
  \label{fig:2}
\end{figure}

\begin{figure}[t]
  \centering
  \includegraphics[width=0.7\linewidth]{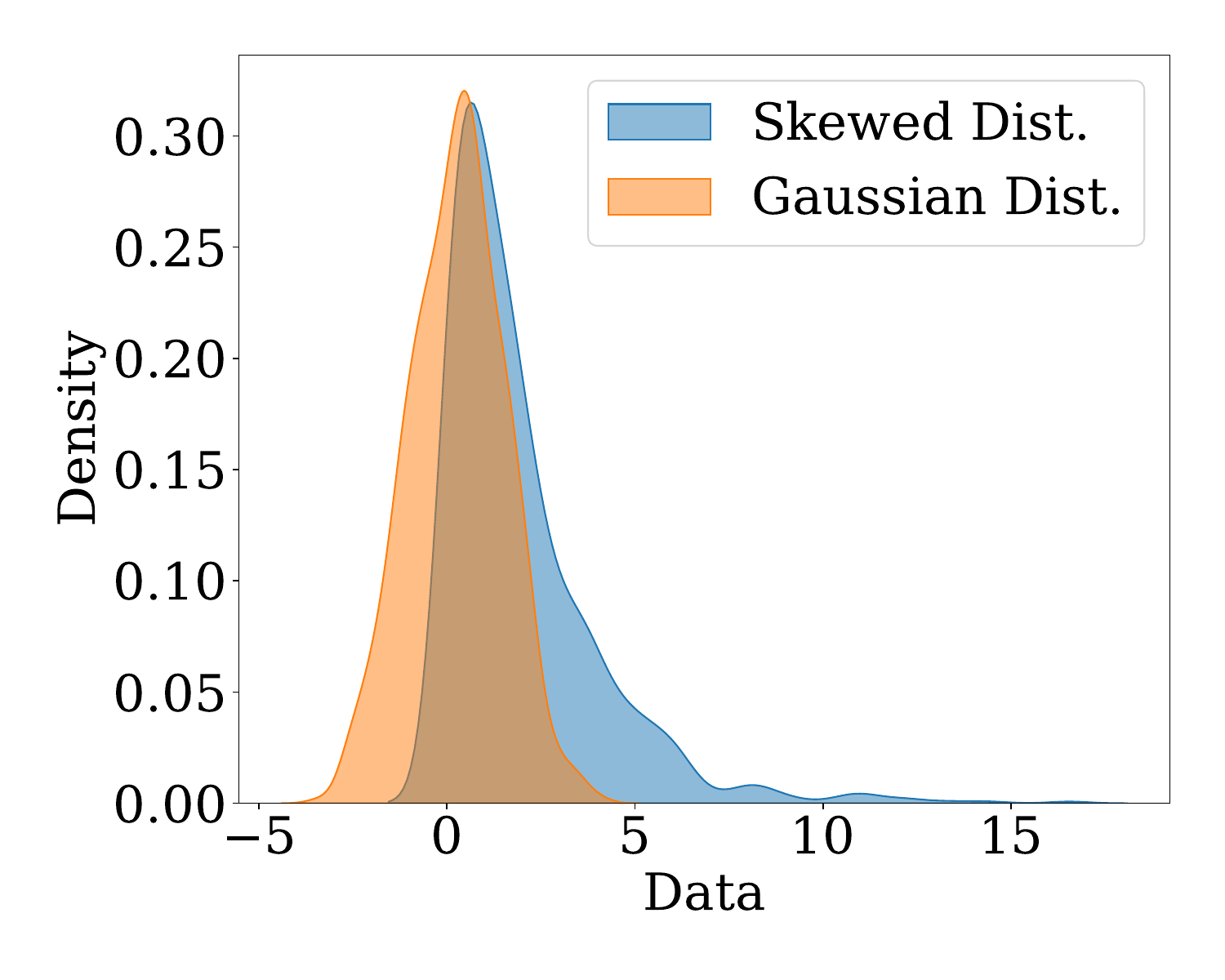}
  \caption{Application of the Box-Cox transformation to normalize a skewed distribution, resulting in a Gaussian-like distribution.}
  \label{fig:3}
\end{figure}

To achieve our objective, it is essential to mitigate extreme alterations in the distribution through appropriate weighting. In this process, while preserving the tail data in its respective position, the anomaly score distribution must exhibit a light-tailed behavior, as suggested by preliminary results. To accomplish this, we must carefully adjust the weighting scheme to ensure that the anomaly score distribution conforms to a light-tailed normal distribution. This objective presents two key challenges: (1) How can the weights be adjusted to shape the anomaly score distribution into a normal form? (2) How can the anomaly scores be transformed into a normal distribution while maintaining their inherent ranking, ensuring that tail data remain in the tail? To address these challenges, we leverage importance sampling in conjunction with the Box-Cox transformation.

Importance sampling is employed when the direct estimation of a probability distribution $p(x)$ is challenging, whereas an alternative distribution $q(x)$ can be more easily obtained. The fundamental formulation of importance sampling is expressed as follows:
\begin{equation}
    \begin{split}
        \mathbb{E}_{p(x)}[f(x)]&=\int f(x)p(x) dx\\
        &=\int f(x)p(x)\frac{q(x)}{q(x)} dx\\
        &=\mathbb{E}_{q(x)}[f(x)\frac{p(x)}{q(x)}].        
    \end{split}
\end{equation}
Note that we aim to make the distribution follows a normal distribution. Therefore, we define the probability $p(x)$ as a normal distribution and $q(x)$ as the distribution of anomaly scores. By utilizing importance sampling as a weighting mechanism, we can effectively reshape the anomaly score distribution to approximate a Gaussian distribution. However, this approach poses the next challenge: how to extract $p(x)$ from a normal distribution while preserving the order of anomaly scores. We employ the Box-Cox transformation, which effectively converts a skewed distribution into a Gaussian distribution, as illustrated in Fig.~\ref{fig:3}. The Box-Cox transformation is formulated as follows:
\begin{equation}
    \begin{split}
        f_b(s_i) &= \begin{cases}
            \frac{s^\lambda_i-1}{\lambda}, & \text{if $\lambda \neq 0$},\\
            \text{ln}(s_i), & \text{if $\lambda = 0$},
        \end{cases}
        \quad \textnormal{for} \quad s_i >0,
    \end{split}    
    \label{eq:boxcox}
\end{equation}
where $s_i$ is an anomaly score and $\lambda$ is a parameter that determines the exponent to which a variable is raised. The optimal value of $\lambda$ is determined that results in the best approximation of a normal distribution curve. By applying this transformation, we obtain a Gaussian distribution derived from the anomaly scores while preserving the relative ranking of the scores. Leveraging importance sampling weights computed from both the original anomaly score distribution and its transformed Gaussian counterpart, we enable the model to learn a distribution that is devoid of long-tailed characteristics, thereby improving its robustness and anomaly detection performance.

Algorithm~\ref{alg:main} provides a summary of the proposed loss function. First, anomaly scores are computed (Line 5). Subsequently, to determine the upper bound $T$ of the weight vector $\textbf{w}$, the skewness of the anomaly scores $f_s(\textbf{s})$ is calculated (Line 6) using the formula  
\begin{equation}
    f_s(\textbf{s})=\frac{1}{N}\sum^N_{i=1}(\frac{s_i-\mu}{\sigma})^3.
    \label{eq:skew}
\end{equation}
Here, $\mu$ and $\sigma$ represent the mean and standard deviation of the anomaly scores, respectively. To ensure that the transformed distribution retains characteristics similar to the original distribution, we first scale the anomaly scores to satisfy the subject of the Box-Cox transformation (Line 7). Next, we apply the Box-Cox transformation (Line 8) to normalize the distribution. For probability estimation, we utilize Gaussian estimation instead of kernel density estimation (KDE), as KDE tends to overestimate probabilities relative to the true distribution (Lines 9-10). However, conventional Gaussian estimation is susceptible to the influence of outliers. To mitigate this issue, we incorporate a modified z-score outlier detection mechanism, formulated as follows.
\begin{equation}
    \begin{split}
        f_p(\textbf{s}) &=\frac{1}{\sqrt{2\pi \sigma^2}}\exp{\big(-\frac{(\textbf{s}-\mu)^2}{2\sigma^2}\big)},\\
        \mu&=\mathbb{E}[\hat{s}_i],\\
        \sigma^2&=\text{Var}(\hat{s}_i),\\
        \hat{\textbf{s}}&=\{s_i|s_i\leq t\},
    \end{split}
    \label{eq:gde}
\end{equation}
where $\mu$ and $\sigma$ denote the mean and standard deviation of the scores in a set $\hat{\textbf{s}}$, respectively. Elements of the set $\hat{\textbf{s}}$ represent the anomaly scores after the removal of outliers, while $t$ is a threshold used to identify outliers. Using a modified z-score outlier detector, the threshold $t$ is determined by
\begin{equation}
    \begin{split}
        t&=3.5*\textnormal{MAD}/0.6745+\hat{s},\\
        \text{MAD} &= \text{median}_{i \in \{1, \cdots,N\}} (|s_i-\bar{s}|), \\
        \bar{s} &= \text{median}_{i \in \{1,\cdots,N\}} (s_i).   
    \end{split}
\end{equation}
Here, MAD is a median absolute deviation and $\bar{s}$ is a median of elements in $\textbf{s}$. After that, we compute the importance sampling weights based on the probabilities derived from both the original and transformed distributions (Line 11).

To ensure stable convergence and mitigate the issue of exploding gradients caused by excessively high weights, we impose an upper bound on the weights. This upper bound is dynamically determined based on the skewness of the distribution and further constrained by a hard limit, $T_0$ (Lines 12-13). Finally, using these weights, we train the model 
$f$ with a weighted loss function (Lines 14–15).

\begin{algorithm}[t]
   \caption{Long-tailed Distribution Loss for Anomaly Detection}
   \label{alg:main}
    \begin{algorithmic}[1]
       \State  {\bfseries Input:} Sample $X$, model $f$, hyperparameters, $\alpha$ and $T_0$
       \State $\epsilon=0.0001$
       \For{\textbf{each} epoch}
           \For{\textbf{each} Mini-batch $\bf{x} \subseteq \bf{X}$}
               \State $\textbf{s}=||\textbf{x}-f(\textbf{x})||^2_2$ \hfill // Calculate anomaly scores
                \State $T=\text{max}(|f_s(\textbf{s})|,\epsilon)$ \hfill // Calculate skewness \eqref{eq:skew}
                \State $\textbf{s}=\textbf{s}-\text{min}(\textbf{s})+\epsilon$
                \State $\textbf{b}=f_b(\textbf{s})$ \hfill // Box-Cox Transformation~\eqref{eq:boxcox}

                \State $\textbf{p}_s = f_p(\textbf{s})$ \hfill // Estimate the probability of $\textbf{s}~\eqref{eq:gde}$
                \State $\textbf{p}_b = f_p(\textbf{b})$ \hfill // Estimate the probability of $\textbf{b}~\eqref{eq:gde}$

                \State $\textbf{w} = \textbf{p}_b/\textbf{p}_s$ \hfill // Importance sampling weight
                \State $T=\text{min}(\alpha T, T_0)$ \hfill //Calculate upper-bound
                \State $\textbf{w} = \text{clip}(\textbf{w},0,T)$
               
                \State $L = \frac{1}{N}\sum^N_{i=1}w_i * ||x_i-f(x_i)||^2_2$ 
                \State Update model parameters with $L$
           \EndFor
       \EndFor
    \end{algorithmic}
\end{algorithm}

\begin{figure}[t]
  \centering
  \includegraphics[width=0.7\linewidth]{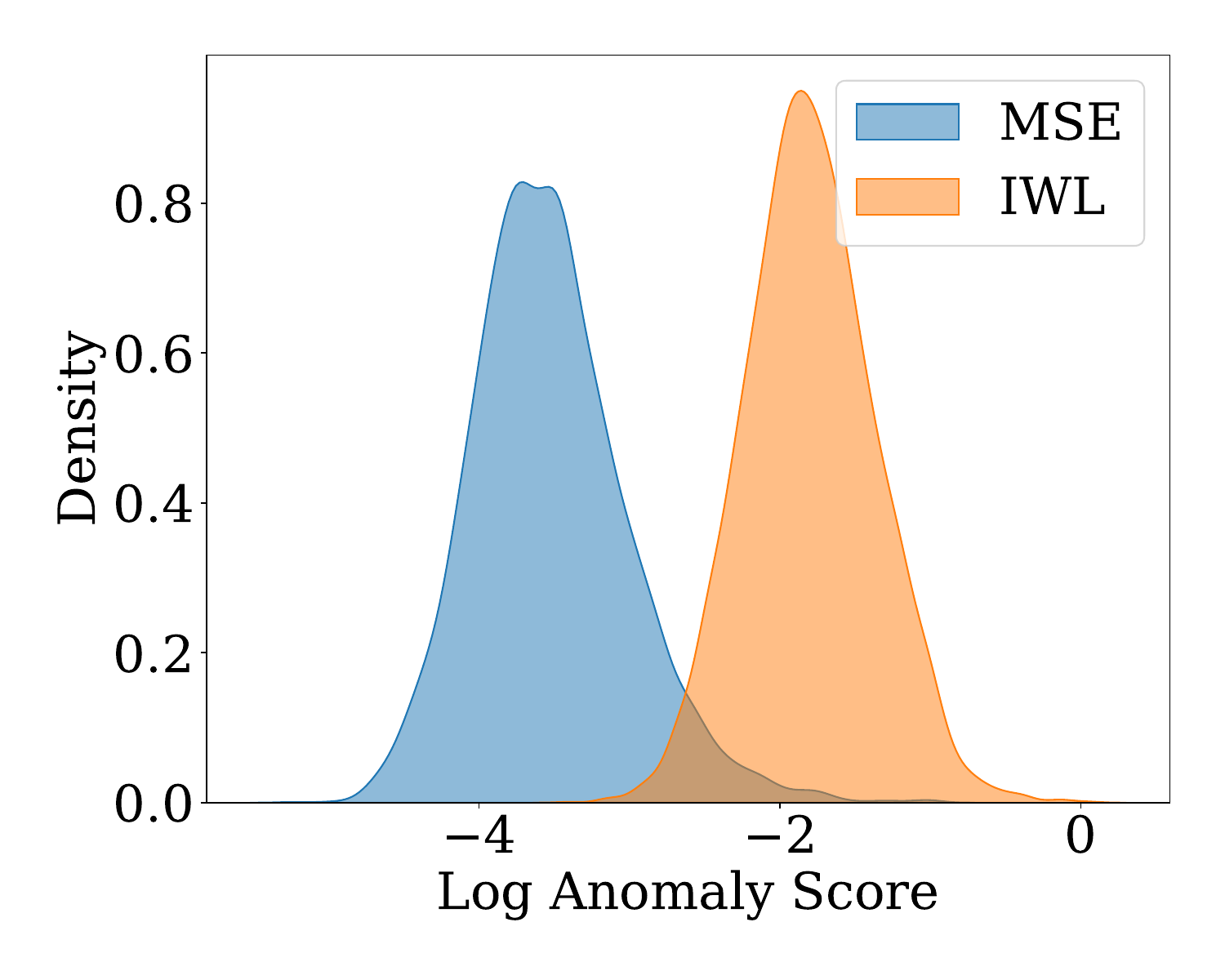}
  \caption{Visualize the distribution of log-scaled scores on the MNIST training dataset using DSVDD.}
  \label{fig:4}
\end{figure}

\begin{table*}[t]
\caption{Evaluation on three long-tailed datasets with $\beta=200$. The highest performance values are highlighted in bold.}
\label{tab:1}
\begin{center}
\resizebox{\linewidth}{!}{%
    \begin{tabular}{|c|c|c|c|c|c|c|c|c|c|c|c|c|c|}
    
    \hline
    \multirow{2}{*}{\textbf{Model}} & \multirow{2}{*}{\textbf{Dataset}} & \multirow{2}{*}{\textbf{Loss}} & \multicolumn{10}{c}{\textbf{Majority Class}}&\\
    \cline{4-14}
    &&& \textbf{0} & \textbf{1} & \textbf{2} & \textbf{3} & \textbf{4} & \textbf{5} & \textbf{6} & \textbf{7} & \textbf{8} & \textbf{9} & \textbf{Avg.}\\
    \hline
    \multirow{6}{*}{AE}&\multirow{2}{*}{MNIST}
    & MSE & 0.976 & 0.797 & 0.710 & 0.657 & 0.651 & 0.682 & 0.712 & 0.647 & \textbf{0.687} & 0.625 & 0.714 \\
    && IWL & \textbf{0.979} & \textbf{0.805} & \textbf{0.715} & \textbf{0.662} & \textbf{0.658} & \textbf{0.689} & \textbf{0.737} & \textbf{0.647} & 0.682 & \textbf{0.654 }& \textbf{0.723} \\
    \cline{2-14}
    
    &\multirow{2}{*}{F-MNIST}
    & MSE & 0.719 & 0.798 & 0.752 & 0.783 & \textbf{0.575} & \textbf{0.628} & 0.559 & 0.766 & 0.659 & 0.721 & 0.696 \\
    && IWL & \textbf{0.730} & \textbf{0.811} & \textbf{0.759} & \textbf{0.789} & 0.558 & 0.626 & \textbf{0.572} & \textbf{0.779} & \textbf{0.664} & \textbf{0.764} & \textbf{0.705} \\
    \cline{2-14}
    
    &\multirow{2}{*}{CIFAR10}
    & MSE & \textbf{0.470} & \textbf{0.460} & 0.610 & \textbf{0.622} & 0.639 & 0.553 & 0.516 & \textbf{0.506} & \textbf{0.520} & \textbf{0.465} & 0.536 \\
    && IWL & 0.466 & \textbf{0.460} & \textbf{0.612} & \textbf{0.622} & \textbf{0.644} & \textbf{0.560} & \textbf{0.531} & 0.501 & 0.516 & 0.458 & \textbf{0.537} \\ 
    \hline

    \multirow{6}{*}{DSVDD}&\multirow{2}{*}{MNIST}
    & MSE & 0.849 & 0.790 & 0.730 & 0.655 & 0.624 & 0.653 & 0.717 & 0.722 & 0.796 & 0.712 & 0.725 \\
    && IWL & \textbf{0.880} & \textbf{0.832} & \textbf{0.772} & \textbf{0.666} & \textbf{0.705} & \textbf{0.723} & \textbf{0.774} & \textbf{0.761} & \textbf{0.805} & \textbf{0.763} & \textbf{0.768} \\
    
    \cline{2-14}
    
    &\multirow{2}{*}{F-MNIST}
    & MSE & \textbf{0.807} & 0.824 & 0.755 & \textbf{0.833} & 0.651 & \textbf{0.669} & \textbf{0.595} & 0.858 & 0.682 & 0.797 & 0.747 \\
    && IWL & 0.790 & \textbf{0.835} & \textbf{0.784} & 0.827 & \textbf{0.675} & 0.662 & 0.593 & \textbf{0.868} & \textbf{0.689} & \textbf{0.822} & \textbf{0.754 }\\
    
    \cline{2-14}
    
    &\multirow{2}{*}{CIFAR10}
    & MSE & \textbf{0.508} & \textbf{0.515} & 0.575 & 0.616 & \textbf{0.604} & 0.602 & 0.557 & 0.514 & 0.564 & 0.506 & 0.556 \\
    && IIWL & 0.501 & 0.513 & \textbf{0.577} & \textbf{0.636} & 0.597 & \textbf{0.613} & \textbf{0.568} & \textbf{0.519} & \textbf{0.565} & \textbf{0.535} & \textbf{0.562} \\

    \hline
    
    \end{tabular}
}
\end{center}
\end{table*}

\begin{figure*}[t]
  \centering
  \includegraphics[width=\linewidth]{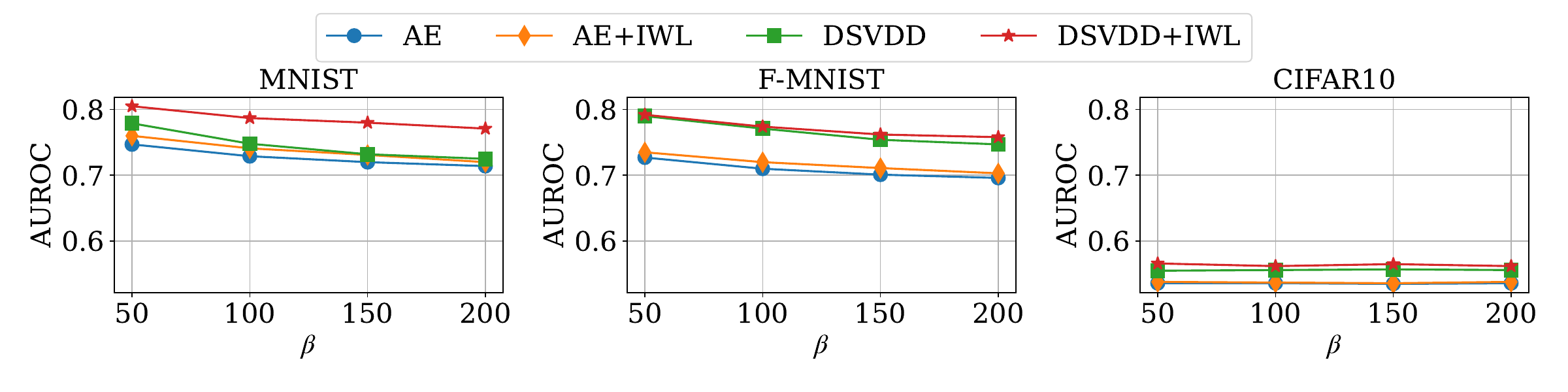}
  \caption{Evaluation across various $\beta$ value with three image datasets.}
  \label{fig:5}
\end{figure*}

\section{Experiments}
\subsection{Datasets} 
We evaluate IWL on three widely used benchmark datasets: MNIST~\cite{lecun2010mnist}, FashionMNIST (F-MNIST)\cite{DBLP:journals/corr/abs-1708-07747}, and CIFAR-10\cite{Krizhevsky09learningmultiple}. In conventional anomaly detection settings, each class is treated as normal while all other classes are considered anomalous~\cite{gong2019memorizing,ruff2018deep}. To simulate a long-tailed distribution, we designate the $i$-th class as the majority class and the neighboring $\text{mod}(i+1,10)$-th class as the minority class. The number of samples in the minority class, $N_{\text{minority}}$, is defined as $\frac{N}{\beta}$, where $N$ represents the number of samples in the majority class and $\beta$ is given by $\frac{N}{N_{\text{minority}}}$. For image datasets, we employ the Area Under the Receiver Operating Characteristic (AUROC) curve as the evaluation metric. Our experimental results report the average AUROC across all ten classes, using three different random seeds, where each class is sequentially considered as the majority class. 

To assess the practical applicability of our approach in an industrial setting, we further evaluate it on three hyperspectral imaging (HSI) datasets~\cite{963e-1d34-24} for food anomaly detection, where the normal objects are almonds, garlic stems, and pistachios. This dataset presents unique challenges, as the normal training data primarily consist of background and target normal objects, introducing substantial spectral variability. Consequently, the dataset exhibits a long-tailed distribution. The dataset consists of 224-band hyperspectral images with a resolution of $512\times400$, with one training image and one test image per category. We use only 10\% of the training data for training. The test data contain rare anomalies, constituting only 8.9\% of the total test set. Given the significant class imbalance, we use the Area Under the Precision-Recall Curve (AUPR) instead of AUROC, as AUPR provides a more reliable assessment of anomaly detection performance in highly imbalanced scenarios.

\subsection{Setups} 
To assess the general applicability of IWL, we evaluate its performance using two representative anomaly detection models: Autoencoder (AE)\cite{hinton2006fast,bergmann2018improving} and Deep Support Vector Data Description (DSVDD)\cite{ruff2018deep}. AE is a reconstruction-based model trained to replicate its input, using reconstruction error as an anomaly detection criterion. In contrast, DSVDD is designed to map normal data to a compact region around a central point in the latent space. 

For image datasets, the architecture of AE adheres to the design proposed by \cite{gong2019memorizing}. The encoder comprises three convolutional modules, each consisting of a convolutional layer, batch normalization~\cite{ioffe2015batch}, and a leaky ReLU activation function~\cite{xu2015empirical}, with filter sizes of 16, 32, and 64, respectively. The kernel and stride sizes are set to 3 and 2, respectively. Similarly, for DSVDD, the autoencoder architecture follows the framework outlined in \cite{ruff2018deep}. The encoder is structured with two convolutional layers containing 8×5×5 and 4×5×5 filters, followed by a fully connected layer with 32 units. Batch normalization, leaky ReLU activation, and (2×2) max-pooling are applied subsequent to the convolutional layers. To mitigate trivial solutions, the biases in the DSVDD layers are eliminated, as recommended in \cite{ruff2018deep}. The decoder is symmetrically designed to mirror the encoder, where deconvolution operations substitute convolutions, and upsampling replaces max-pooling. The final deconvolution layer does not incorporate additional operations, such as batch normalization. For HSI datasets, the encoder architecture of both AE and DSVDD consists of five fully connected layers, each followed by batch normalization and leaky ReLU activation, except for the final layer. Similar to the approach used for image datasets, biases are removed in DSVDD to prevent trivial solutions. Additionally, the decoder is designed symmetrically to the encoder.

For model training, we used 100 epochs across all models. Additionally, pre-training in DSVDD was conducted for 150 epochs on image datasets and 20 epochs on HSI datasets. All models were optimized using the Adam optimizer~\cite{kingma2014adam} with a learning rate of 0.0001 and a weight decay of 
$10^{-6}$. The hyperparameters $\alpha$ and $T_0$ for the upper bound $T$ were set to 4 and 20, respectively.

\subsection{Visualization Distribution Transformation}
Fig.~\ref{fig:4} visualizes the distribution of log-scaled anomaly scores on the MNIST training dataset using DSVDD. The distribution with MSE exhibits a skewness of 0.623, whereas the distribution with IWL has a skewness of 0.176. IWL reduces skewness by 0.447, demonstrating its effectiveness in projecting the anomaly score distribution toward a Gaussian form.

\subsection{Evaluation on Long-Tailed Image Datasets}
We evaluate IWL with conventional AE and DSVDD models, employing Mean Squared Error (MSE) loss as the baseline. Table~\ref{tab:1} presents the performance comparison of AE and DSVDD trained with both MSE and IWL across three datasets, using $\beta=200$. Notably, IWL consistently outperforms MSE, yielding an average AUROC improvement of 0.43. This effect is particularly pronounced in DSVDD on the MNIST dataset, where the data distribution exhibits strong skewness. The performance gains achieved by IWL gradually diminish as the datasets become progressively more complex—from MNIST to CIFAR-10—the anomaly score distribution increasingly approximates a Gaussian distribution. Fig.~\ref{fig:5} illustrates the average AUROC as a function of the number of samples in the minority class. While IWL consistently outperforms MSE, the performance gap between the two methods narrows as $\beta$ decreases. These results demonstrate that IWL effectively mitigates the adverse effects of long-tailed distributions, leading to enhanced anomaly detection performance across diverse datasets.

\subsection{Evaluation on Real-world Datasets}
We evaluate on real-world industrial dataset, which has three hyperspectral imaging (HSI). This dataset is specifically designed for identifying various foreign substances in food products. Unlike standard benchmark datasets, the normal data in this dataset comprises both background elements and diverse spectral variations of raw materials, inherently inducing LTD. This characteristic makes it an ideal dataset for evaluating the effectiveness of our proposed algorithm. The HSI dataset consists of three subsets, each focusing on a distinct primary object: almonds, garlic stems, and pistachios. Table~\ref{tab:2} presents the AUPR results for AE and DSVDD on this dataset. Notably, AE with IWL achieves an improvement of 0.12, while DSVDD with IWL demonstrates a performance gain of 0.033 compared to their respective baselines. These results validate the effectiveness of our approach in enhancing anomaly detection in real-world datasets, highlighting its strong applicability across various industrial domains.

\begin{table}[t]
\caption{Evaluation on real-world hyperspectral anomaly detection datasets with five different seeds. The highest performance values are highlighted in bold.}
\label{tab:2}
\begin{center}
\resizebox{\linewidth}{!}{
\begin{tabular}{|c|c|c|c|c|c|}
\hline
\textbf{Model} & \textbf{Loss} & \textbf{Almond} & \textbf{GarlicStems} & \textbf{Pistachio} & \textbf{Average}\\
\hline
\multirow{2}{*}{AE}
& MSE & 0.716 & 0.885 & \textbf{0.793} & 0.797 \\
\cline{2-6}
& IWL & \textbf{0.728} & \textbf{0.897} & 0.780 & \textbf{0.801}\\

\hline
\multirow{2}{*}{DSVDD}
& MSE& 0.705 & 0.816 & 0.769 & 0.763 \\
\cline{2-6}
& IWL & \textbf{0.738} & \textbf{0.846} & \textbf{0.799} & \textbf{0.794} \\

\hline
\end{tabular}}
\end{center}
\end{table}

Table~\ref{tab:3} presents a comparative analysis of our approach against state-of-the-art anomaly detection models. In addition to AE and DSVDD, we evaluate several recent anomaly detection methods, including Reconstruction Along Projection Pathway (RAPP)\cite{Kim2020RaPP}, Memory-Augmented Autoencoder (MemAE)\cite{gong2019memorizing}, DIF~\cite{xu2023deep}, and SLAD~\cite{xu2023fascinating}. The experimental results reveal that existing models struggle to effectively learn from long-tailed data, often misclassifying normal samples as anomalies, which leads to substantial performance degradation. In contrast, IWL successfully mitigates the challenges posed by long-tailed distributions, significantly enhancing the performance of its base models. These findings demonstrate that IWL not only improves anomaly detection by addressing the limitations associated with long-tailed distributions but also serves as a robust and generalizable solution applicable across various anomaly detection frameworks.

\begin{table}[t]
\caption{Evaluation with state-of-the-art methods on three real-world datasets across five different seeds. The highest performance values are highlighted in bold.}
\label{tab:3}
\begin{center}
\begin{tabular}{|c|c|c|c|c|}
\hline
\textbf{Model}&\textbf{Almond} & \textbf{GarlicStems} & \textbf{Pistachio} &\textbf{Average}\\
\hline
        AE & 0.716 & 0.885 & 0.789 & 0.797 \\ \hline
        DSVDD & 0.705 & 0.816 & 0.769 & 0.763 \\ \hline
        RAPP & 0.701&	0.659&	0.674&	0.678 \\ \hline
        MemAE & 0.327 & 0.131 & 0.275 & 0.244 \\ \hline
        DIF & 0.073 & 0.044 & 0.043 & 0.053 \\ \hline
        SLAD & 0.449 & 0.465 & 0.440 & 0.452 \\ \hline
        AE+IWL & 0.728 & \textbf{0.897} & 0.780 & \textbf{0.801} \\ \hline
        DSVDD+IWL & \textbf{0.738} & 0.846 &\textbf{0.799} & 0.794 \\ 
\hline
\end{tabular}
\end{center}
\end{table}

\section{Conclusions}
In this study, we propose the Importance-Weighted Loss (IWL) function to address the real-world challenge of long-tailed distributions (LTD) in anomaly detection. Our approach is grounded in the theoretical assumption that anomaly scores should approximate a Gaussian distribution. Based on this premise, we apply the Box-Cox transformation to induce a Gaussian-like distribution and leverage importance sampling to enhance anomaly detection performance. Extensive experiments conducted on three benchmark image datasets and three hyperspectral imaging datasets demonstrate that IWL effectively improves the performance of AE and DSVDD by 0.043, enabling them to better handle long-tailed distributions compared to conventional methods. While IWL successfully mitigates the LTD problem, our current study does not explicitly focus on optimizing the training process or enhancing model generalization which we plan to explore in future work. In addition, we aim to integrate data augmentation techniques and contrastive learning to further improve performance. Beyond these enhancements, we intend to evaluate IWL’s effectiveness in addressing LTD-related challenges in diverse applications, such as medical anomaly detection and financial fraud detection. In conclusion, IWL effectively alleviates underfitting issues induced by LTD, significantly enhancing anomaly detection performance. Although this method requires only a modification of the loss function without altering the underlying model architecture, it offers a practical and efficient solution for real-world industrial applications.




\vspace{12pt}

\end{document}